\documentclass{article} 
\usepackage{iclr2017_workshop,times}
\usepackage{hyperref}
\usepackage{url}
\usepackage{amsmath,amssymb}
\usepackage{hyperref}
\usepackage{amsfonts}
\usepackage{graphicx}
\usepackage{booktabs}
\usepackage{subcaption}
\usepackage{float}
\usepackage{comment}

\title{Efficient variational Bayesian neural network ensembles for outlier detection}

\author{Nick Pawlowski\thanks{Equal contribution.}\\
Imperial College London\\
\texttt{np716@ic.ac.uk} \\
\And
Miguel Jaques\footnotemark[1]\\
No affiliation \\
\texttt{migjaques@gmail.com}
\And
Ben Glocker\\
Imperial College London \\
\texttt{b.glocker@ic.ac.uk}
}

\begin{document}
\maketitle
\vspace{-0.4cm}
\begin{abstract}
In this work we perform outlier detection using ensembles of neural networks obtained by variational approximation of the posterior in a Bayesian neural network setting. The variational parameters are obtained by sampling from the true posterior by gradient descent. We show our outlier detection results are comparable to those obtained using other efficient ensembling methods.
\end{abstract}

\vspace{-0.4cm}

\section{Introduction and related work}
In this work we use posterior sampling by gradient descent in order to estimate the parameters of a variational approximation of the true posterior. This allows us to create an ensemble of networks that we use to perform outlier detection. We show that our method achieves better results than other efficient ensembling methods, though worse than standard ensembles.
Our work is inspired by a mix of scalable ways to perform Bayesian deep learning (\cite{graves2011practical,blundell2015weight,hernandez2015probabilistic,gal2015dropout,mcclure2016representation}), the use of ensembles to improve accuracy and obtain uncertainty estimates of the network's predictions
(\cite{zhou2002ensembling, lakshminarayanan2016simple}), and the ability to sample from the true posterior using stochastic gradient descent-like techniques (\cite{welling2011bayesian,teh2016consistency}). We use these  samples to obtain a variational approximation of the true posterior, which can then be used to generate an ensemble of networks. We take inspiration from \cite{huang2016snapshot}, where weight samples are taken along the optimization trajectory, and used as an ensemble at test time. Instead, we take weight samples and use them to incrementally estimate the parameters of a variational approximation to the posterior during training, which allows us to sample from a distribution over weights while keeping constant memory requirements.

In \cite{lakshminarayanan2016simple}, the authors train $N$ independent networks from scratch (with respective uncertainty estimates for regression) and use the ensemble to obtain good uncertainty estimates and classification performance. We view this model as an upper bound on how good our model can be, since the goal is to achieve similarly good results training only a single network.

This work is most comparable to \textit{Dropout-MC} (\cite{gal2015dropout}) and other Dropout-related methods (\cite{mcclure2016representation}). This is because only a single network is trained and the ensembles are only created for testing.

\section{Variational Bayes}
In Bayesian inference we are interested in evaluating the predictive distribution
\begin{equation}
p(y|x,D) = \int p(y|x,w) p(w|D) dw \label{eq:predictive}
\end{equation}
where $p(w|D) = p(D|w)p(w)/p(D)$ is the posterior over the weights, $D = {(x^i, y^i)}_{i=1}^M$ is the training data, $p(D|w)$ is the model likelihood, parametrized by a neural network with e.g. a softmax or Gaussian output, and $p(w)$ is the prior. In order to evaluate the integral above we first find an approximate posterior $q_{\theta}(w)$, where $\theta$ are the parameters of $q$ (for example, if $q$ is Gaussian, $\theta = (\mu, \sigma)$) and then use use Monte-Carlo samples from this posterior to obtain an ensemble that can be used to estimate (\ref{eq:predictive}). In variational approximations the optimal parameters $\theta^*$ are usually found by minimizing the KL-divergence between the approximate and the true posterior.
However, this form underestimates the variance of the posterior, hence its uncertainty. Since we are interested in a robust estimate of the uncertainty, we instead minimize the forward KL-divergence:
\begin{equation}
\theta^* = \textnormal{argmin}_{\theta} KL(p(w| D)||q_{\theta}(w) ) = \textnormal{argmin}_{\theta} \mathbb{E}_{p(w|D)} \left[-  \log q_{\theta}(w) \right]
\end{equation}

Using a diagonal Gaussian as approximate posterior, $\log q_{\theta}(w) = \sum_i \log q_{\theta_i}(w_i) = \sum_i -\log \sigma_i -\frac{1}{2\sigma_i^2} (w_i - \mu_i)^2 + D$,
where $D$ is a constant, the minimization with respect to $\mu$ and $\sigma^2$ yields
\begin{equation}
\mu^* = \mathbb{E}_{p(w|D)} [w] \quad , \quad
{\sigma^*}^2 = \mathbb{E}_{p(w|D)} [(w-\mu)^2]
\end{equation} 

This shows the optimal approximate weight means and variances are simply the first and second moments of the weights over the posterior distribution.
In order to sample from the intractable posterior distribution of the weights without additional computation or memory requirements besides the standard stochastic gradient descent calculations, we can use Stochastic Gradient Lagevin Dynamics (SGLD) (\cite{welling2011bayesian,teh2016consistency}). We use the trajectory of the weights through parameter space during training as samples from the unnormalized posterior $p(D|w)p(w)$, and  compute $\mu$ and $\sigma^2$ using the standard unbiased Monte-Carlo estimates 
$\hat{\mu} = \sum_s^S w^{(s)}$/S and  
$\hat{\sigma}^2 = \sum_s^S (w^{(s)} - \hat{\mu})^2$ / (S-1),
for $w^{(s)} \sim p(D|w)p(w)/Z $. Due to the large number of parameters in a neural network, storing all the samples (one per mini-batch step) would be too memory expensive. Instead we use the incremental formulas for computing $\hat{\mu}$ and $\hat{\sigma}^2$ from \cite{welford1962note}. This incremental estimation reduces the memory requirements to a minimum since we only have to store a single set of parameters during training. We can use this method as long as our training trajectory works as a valid sampling procedure, i.e, $w^{(s)} \sim p(D|w)p(w)/Z$, which is guaranteed by SGLD.

\section{Ensembles and outlier detection}
At test time, we sample weights from $q_{\theta}(w)$ in order to obtain a neural network ensemble and calculate the predictive probability using a Monte-Carlo estimate of \eqref{eq:predictive}. The ensemble is used to calculate an uncertainty score of the prediction. We use \textit{disagreement} as defined by \cite{lakshminarayanan2016simple} as uncertainty measure. The disagreement is defined as  the sum of the KL-divergence between the average ensemble prediction and each ensemble component. We use the calculated disagreement to perform outlier detection. Therefore, we calculate the mean $\mu(d_{train})$ and standard deviation $\sigma(d_{train})$ of the disagreement $d_{train}$ for a random training batch. A given sample is labelled as an outlier by thresholding:
\begin{align}
p(\mathrm{outlier}\mid x) = d_{x} \geq \mu(d_{train}) + 3\sigma(d_{train})
\end{align}

Here, the addition of $3\sigma(d_{train})$ to the threshold is used to minimize the false positive rate.

\section{Experiments and results}
In the experiments we used a 3-layer fully-connected neural network with 200 hidden units per layer and 10-unit softmax output, and a standard Gaussian prior. We trained on the MNIST dataset and run outlier detection on the MNIST (\cite{lecun1998mnist}) and notMNIST (\cite{bulatov2011notmnist}) test sets. The networks were trained with a batch size of 100 for 1000 steps. We compare our method against Dropout-MC (\cite{gal2015dropout}) and standard ensembles (\cite{lakshminarayanan2016simple}) where each network is trained from scratch. Experiments with Snapshot Ensembles (\cite{huang2016snapshot}) and fully factorized variational inference were run but did not yield good performances.

SGLD imposes some conditions on the step sizes and gradient noise (such as annealing) in order to guarantee that our procedure is a valid sampler from the true posterior. Because of this we compare our method to Dropout-MC and standard ensembles trained with the same learning rate schedule. Additionally, we introduce an intuitive approximation to SGLD using an Adam optimizer (\cite{kingma2014adam}) with added noise to the updates. The noise was scaled according to the adaptive learning rate calculated by Adam. We compare this method to Dropout-MC and standard ensembles trained with Adam. The results can be found in the appendix. All experiments\footnote{The code is available at \url{https://github.com/pawni/sgld_online_approximation}.} use a base learning rate of $0.005$. We do not use any burn in or thinning when calculating the Gaussian approximation.

Table \ref{ensemble-accuracy} shows the classification accuracies for the different models using the SGLD learning rate schedule given the number of ensemble components.  The classification performance of our method is situated between Dropout-MC and standard ensembles. However, the performance of our method does not increase with the number of components in the ensemble. Analysing the disagreement results across the ensembles shows very small disagreement. Our method shows disagreements of order $10^{-3}$ to $10^{-2}$, whereas the other methods have a disagreement of $10^{-1}$ to $10^{0}$. This means that the variational approximation fits a Gaussian with small variance. However, outlier detection depends on relative rather than absolute disagreement values between in-dataset samples and out-of-dataset samples.

\begin{table}
\centering
\caption{MNIST classification accuracy of different methods with various ensemble sizes using the SGLD learning rate schedule. Standard ensembles perform best. Our method outperforms Dropout-ensembles but does not improve with higher ensemble size.}
\label{ensemble-accuracy}
\begin{tabular}{l|rrr}
\toprule
Ensemble size & Standard & Dropout & Ours \\ \midrule
1             & $80.7 \pm 1.1$ & $61.4 \pm 3.2$ & $76.0 \pm 0.6$ \\ 
5             & $86.3 \pm 0.6$  & $71.4 \pm 2.9$& $76.1 \pm 0.7$ \\ 
10            & $87.0 \pm 0.6$ & $72.9 \pm 2.7$ & $76.1 \pm 0.7$ \\ \bottomrule
\end{tabular}
\end{table}

\begin{figure}[h]
  \begin{subfigure}[t]{0.45\textwidth}
  \centering
    \includegraphics[width=1.0\textwidth]{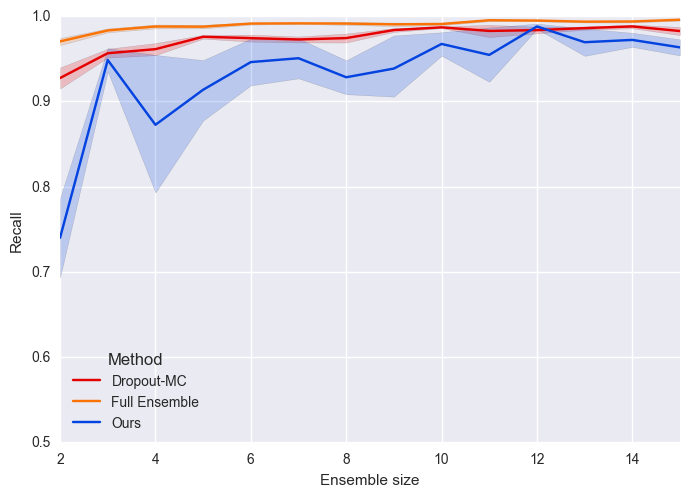}
    \label{fig:true_positive}
    \end{subfigure}
    \hfill
    \begin{subfigure}[t]{0.45\textwidth}
     \centering
     \includegraphics[width=1.0\textwidth]{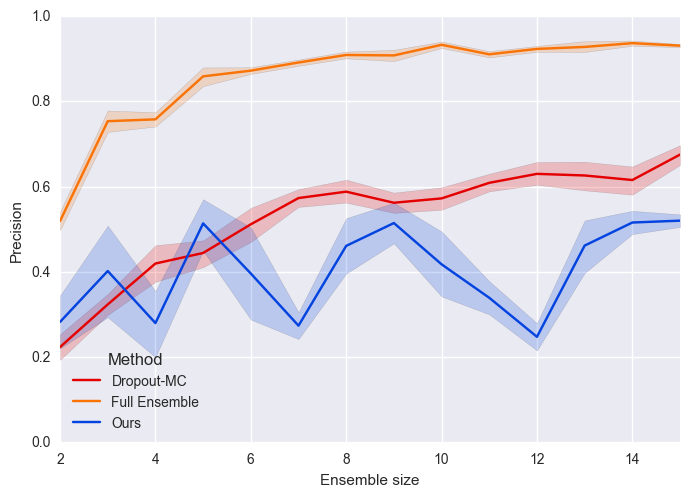}
     \label{fig:false_positive}
     \end{subfigure}
     \caption{Recall and precision for outlier detection using different ensembling methods and various ensemble sizes. The methods are tested on the test sets of MNIST and notMNIST (outlier). All methods show high recall. Standard ensembles show high precision but Dropout-MC and our method have lower precision. Our method performs slightly worse than Dropout-MC.}
     \label{fig:outlier}
\end{figure}

We perform outlier detection on the test set of both MNIST and notMNIST.  The results for various ensemble sizes using the SGLD learning rate schedule can be seen in Fig. \ref{fig:outlier}. Standard ensembles achieve the best overall performance. This can be explained by the use of different initialisations which lead to a higher variety in trained models. Our method method and Dropout-MC achieve high recall but low precision. Our method performs slightly worse than Dropout but outperforms it on the the normal MNIST classification task. Additional results using Adam are shown in the appendix.

\section{Conclusion}
We present an efficient method for building variational approximations of neural networks. It enables the creation of ensembles. Our method is easily adaptable to a wide range of neural network architectures, provides comparable prediction performance and only minimally increases computation and memory requirements during training. Furthermore, this method can be combined with other sampling methods and a combination with recent advances in gradient-based approximate Bayesian inference \cite{mandt2017stochastic} is left for future work. Additionally, we propose a simple ensemble based outlier detection method which is inspired by \cite{lakshminarayanan2016simple}.

\subsubsection*{Acknowledgements}
NP is supported by Microsoft Research through its PhD Scholarship Programme and the EPSRC Centre for Doctoral Training in High Performance Embedded and Distributed Systems  (HiPEDS, Grant Reference EP/L016796/1).

\bibliographystyle{iclr2017_workshop}
\bibliography{biblio}
\pagebreak
\appendix
\section{Results using \textit{noisy Adam} as approximation for SGLD}
Here we present the results obtained by using \textit{noisy Adam} as approximation for SGLD. We define the \textit{noisy Adam} update as 
\begin{align*}
\Delta\theta = \text{Adam}(\theta, x, y) + \epsilon\quad\mathrm{with}\quad\epsilon \sim \mathcal{N}(0, \alpha_t^2)
\end{align*}
where $\Delta\theta$ is the update of  $\theta$, $\text{Adam}(\theta, x, y)$ is the update of $\theta$ given the data points  $x$ and $y$ according to regular Adam and $\epsilon$ is the noise added to this update. This noise is Gaussian distributed with a standard deviation of  $\alpha_t$ which is the current adaptive learning rate calculated by Adam. Table \ref{ensemble-accuracy-adam} shows the MNIST classification accuracy of different methods with various ensemble sizes using Adam for optimization. Standard ensembles achieve the best performance. Our method outperforms Dropout and is able to achieve a better performance for the single network `ensemble' case.

\begin{table}[h]
\centering
\caption{MNIST classification accuracy of different methods with various ensemble sizes using Adam. Standard ensembles perform best. Our method performs better then Dropout-ensembles and only slightly worse than standard ensembles.}
\label{ensemble-accuracy-adam}
\begin{tabular}{l|rrr}
\toprule
Ensemble size & Standard & Dropout & Ours \\ \midrule
1             & $92.7 \pm 0.5$ & $89.6 \pm 0.8$ & $92.8 \pm 0.9$ \\ 
5             & $94.6 \pm 0.2$ & $91.5 \pm 0.7$& $93.9 \pm 0.2$ \\ 
10            & $94.9 \pm 0.2$ & $91.7 \pm 0.6$ & $94.1 \pm 0.1$ \\ \bottomrule
\end{tabular}
\end{table}

\begin{figure}[h]
  \begin{subfigure}[t]{0.45\textwidth}
  \centering
    \includegraphics[width=1.0\textwidth]{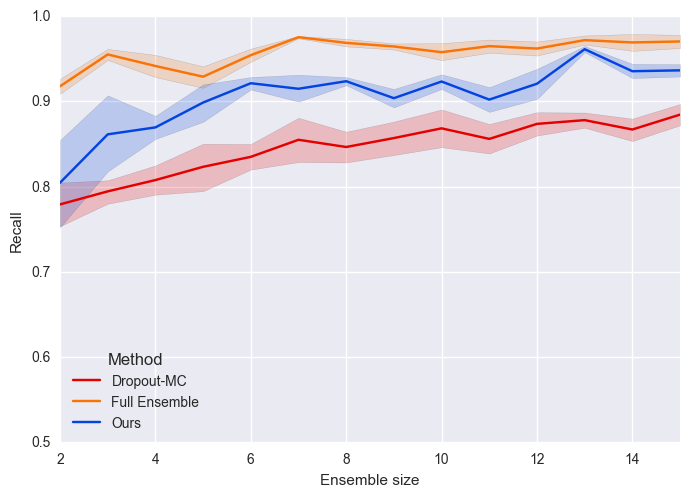}
    \label{fig:adam_recall}
    \end{subfigure}
    \hfill
    \begin{subfigure}[t]{0.45\textwidth}
     \centering
     \includegraphics[width=1.0\textwidth]{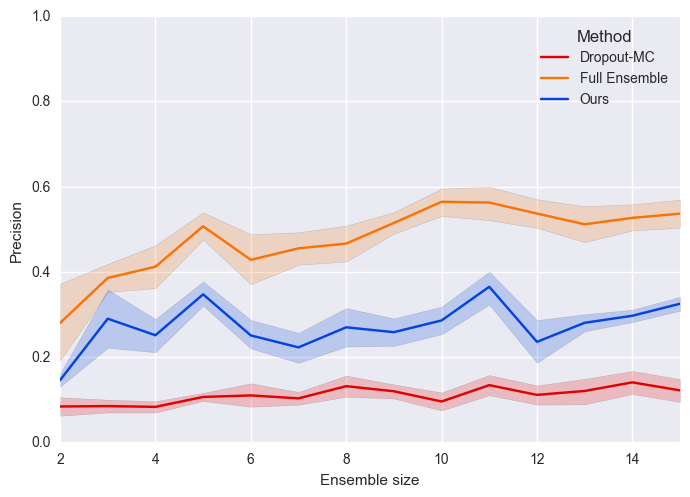}
     \label{fig:adam_precision}
     \end{subfigure}
     \caption{Recall and precision for outlier detection using different ensembling methods and various ensemble sizes. The methods are tested on the test sets of MNIST and notMNIST (outlier). All methods show low precision and high recall. Our method is only outperformed by standard ensembles.}
     \label{fig:outlier_adam}
\end{figure}

We then performed the same outlier detection task with those methods. The results are shown in \autoref{fig:outlier_adam}. Again, standard ensembles perform best. However, \textit{noisy Adam} outperforms Dropout-MC on both recall and precision. Additionally, the general outlier detection performance using Adam is worse than the results using the SGLD learning rate schedule. 

\end{document}